\documentclass[runningheads]{llncs}
\usepackage[T1]{fontenc}
\usepackage{graphicx}
\usepackage{amsmath,amssymb}
\usepackage{hyperref}
\usepackage{booktabs}
\usepackage{float}
\usepackage{url}
\usepackage{bbding}
\begin{document}
\title{On the Definition of Intelligence}

\author{Kei-Sing Ng}
\authorrunning{K.-S. Ng}
\institute{
\email{max.ksng.contact@gmail.com}}

\maketitle

\begin{abstract}
To engineer AGI, we should first capture the essence of intelligence in a species-agnostic form that can be evaluated, while being sufficiently general to encompass diverse paradigms of intelligent behavior, including reinforcement learning, generative models, classification, analogical reasoning, and goal-directed decision-making. We propose a general criterion based on \textit{entity fidelity}: Intelligence is the ability, given entities exemplifying a concept, to generate entities exemplifying the same concept. We formalise this intuition as \(\varepsilon\)-concept intelligence: it is \(\varepsilon\)-intelligent with respect to a concept if no chosen admissible distinguisher can separate generated entities from original entities beyond tolerance \(\varepsilon\). We present the formal framework, outline empirical protocols, and discuss implications for evaluation, safety, and generalization. By defining intelligence based on the principle of generative fidelity to a concept, our definition provides a single yardstick for comparing biological, artificial, and hybrid systems, and invites further theoretical refinement and empirical validation.
\end{abstract}

\section{Introduction}
The sprint toward AGI forces us to confront an old question with renewed urgency: what is intelligence? The concept of intelligence remains difficult to define formally. As artificial systems begin to outperform humans in specialized tasks, and as general-purpose models such as large language models (LLMs)\cite{Brown2020,Radford2019} become increasingly prominent, the need for a clear, operational, and testable definition of intelligence becomes more urgent.

This paper proposes a minimal and foundational definition that seeks to unify human, artificial, and biological intelligence within a single operational framework across generative intelligence\cite{Goodfellow2014,Kingma2013,Heusel2017}, classification\cite{Bishop2006,Duda2001}, reinforcement learning\cite{Sutton1998} and analogical reasoning\cite{gentner1997structure}. It is intended as a theoretical baseline from which future refinements can be developed.

\subsection{Motivation}
We aim to identify a deep and universal characteristic across diverse paradigms of intelligent behavior and to adopt it as a working definition of intelligence. A more robust definition should abstract away from specific tasks, goals, or species and instead capture the fundamental capacity underlying all forms of intelligent behavior. Legg and Hutter surveyed numerous definitions of intelligence\cite{legg2007collection} and proposed that ``intelligence measures an agent’s ability to achieve goals in a wide range of environments.''\cite{legg2007universal} This informal definition effectively captures a general property of reinforcement learning and goal-directed decision-making. However, its applicability becomes less clear when considering intelligent behaviors that lack a well-defined environment, goal, or reward signal. For instance, generative AI systems that produce images or videos often do not operate within a clearly delineated environment-agent framework.

The Turing Test is a historically significant and elegantly designed benchmark for evaluating intelligence, particularly in contexts where the distinction between human and machine responses is subtle.\cite{turing1950computing} However, its applicability is most effective in scenarios where machine intelligence is comparable to, or below, human-level performance. In domains where machine intelligence substantially exceeds human capabilities in specific tasks, alternative evaluation methods may be needed to fully capture such forms of intelligence.

Similar issues arise with definitions such as Elaine Rich's: ``Artificial Intelligence is the study of how to make computers do things at which, at the moment, people are better.''\cite{rich1983artificial} This view is inherently anthropocentric and temporally contingent, as it depends on tasks where humans currently maintain superiority.

From a psychological perspective, David Wechsler defined intelligence as ``the global capacity of the individual to act purposefully, to think rationally, and to deal effectively with his environment.''\cite{wechsler1944measurement} While this definition offers important insights, it is difficult to operationalize, particularly in the context of non-human systems. When attempting to construct an intelligence measure that is not centered on human judgment, new approaches are required.

We boldly propose to adopt a certain property of analogy as the foundational definition of intelligence. Prior research has emphasized the centrality of analogy in cognition\cite{hofstadter2001analogy}. We present this property as an informal definition in Section 2, and demonstrate how key aspects such as creativity and learning ability can be situated within this framework. To formally incorporate relational and structural aspects into the definition, we adopt a precise mathematical notation to capture the similarity between \textit{entities}. In this setting, entities within a concept exhibit a form of similarity, often of an abstract nature. To evaluate such similarity, we define \(\varepsilon\)-concept intelligence using precise mathematical terminology, allowing us to concretely measure intelligence through the notion of indistinguishability, rather than relying on vague descriptions. Ultimately, we provide a new lens that stimulates further discussion and refinement on the nature of intelligence.

\subsection{Desiderata for a Definition}
To be broadly applicable in scientific and engineering contexts, a definition of intelligence should satisfy the following criteria:
\begin{itemize}
    \item \textbf{Operational:} It should support empirical testing of intelligent behavior via observable outputs.
    \item \textbf{Falsifiable:} It should allow clear distinction between intelligent and non-intelligent systems based on measurable deviations.
    \item \textbf{Generalizable:} It should apply uniformly to humans, machines, and natural systems without anthropocentric bias.
\end{itemize}

These criteria align with and simplify a more extensive set of desiderata proposed for intelligence tests, such as those discussed by Hernández-Orallo \cite{HernandezOrallo2000}.

\section{Proposed Definition}

We first offer an informal definition of intelligence:
\begin{quote}
\textit{Intelligence is the ability, given entities exemplifying a concept, to generate entities exemplifying the same concept.}
\end{quote}

Let \(U\) denote the universe of all conceivable entities. Entities are not limited to data points—they can also include products, designs, actions, or even concepts and more. 
A \emph{conceptual space} is a surjective mapping
\[
  \Phi : U \longrightarrow K ,
\]
where \(K\) is the set (or measurable space) of \emph{concepts}, each \(k \in K\) serving as a representation for a specific set of entities in \(U\).
For any concept \(k\in K\) we write
\[
  C_k \;=\; \Phi^{-1}(k) \;\subseteq\; U ,
\]

and call \(C_k\) the \emph{fibre} associated with \(k\).

The given entities form a finite \emph{multiset}\footnote{%
Duplicates are permitted.  Whenever order is irrelevant, we view
$S$ as a finite multiset $S\in M(U)$; when order \emph{is} relevant
(e.g., for sequence-based distinguishers), we instead treat
$S=(x_1,\dots,x_m)\in U^m \text{ as a finite sequence.}$}
\[
  S \;=\; \{x_1,\dots,x_m\}\subseteq U ,
\]
which lies inside some (possibly unknown) fibre \(C_k\).
No probabilistic structure on \(C_k\) is assumed, so even \(m=1\) is permitted.
\medskip

\begin{definition}[$\varepsilon$-Concept Intelligence]
Let \(F\) be a family of distinguishers \(f:U\!\to\![0,1]\).
Let \(S\subseteq U\) be a given entity multiset that is \emph{compatible
with a concept} (i.e., there exists at least one concept \(k \in K\) such that its fibre \(C_k\) contains—or is well-approximated by—\(S\)).
Given any generated entity set \(\widehat{S}\subseteq U\) produced on
the basis of \(S\), and scoring function
$\sigma : M([0,1]) \longrightarrow [0,1]$, define
\begin{equation}
\Delta_{F}^{\sigma}(S,\widehat{S})
   \;=\;
   \sup_{f\in F}
   \bigl|
      \sigma\bigl(\{f(x):x\in S\}\bigr)
      \;-\;
      \sigma\bigl(\{f(x'):x'\in\widehat{S}\}\bigr)
   \bigr|.
\end{equation}
The ability to generate set \(\widehat{S}\) is said to be
\(\varepsilon\)\emph{-intelligent} with respect to the (implicit)
fibre \(C_k\) if \(\Delta_{F}^{\sigma}(S,\widehat{S})\le\varepsilon\).
\end{definition}

\noindent
Informally, \(\widehat{S}\) is indistinguishable from the given
entities \(S\)—up to tolerance \(\varepsilon\)—for every distinguisher in
\(F\), according to the chosen scoring rule \(\sigma\). The principled selection of a concept \(k\) and distinguishers \(F\) is a non-trivial question on which we will elaborate in Section 5.

For example, consider ChatGPT producing Studio Ghibli-style frames: let \(S\) be a set of genuine Ghibli-style images and \(\widehat{S}\) be ChatGPT’s outputs. If human judges (or automated critics) cannot reliably distinguish \(\widehat{S}\) from \(S\) within a small tolerance \(\varepsilon\), then by our definition ChatGPT is \(\varepsilon\)-intelligent in the concept ``Ghibli-style images.'' This scenario aligns with our intuitive notion of ``looks like Ghibli,'' illustrating how indistinguishability underpins both formal measurement and everyday understanding of style‑based intelligence.

Crucially, the similarity among entities in a concept can be highly abstract. For instance, a horse and a car, despite their stark differences in appearance and internal structure, belong to the same concept of ``conveyances for human transport.'' An entity demonstrates intelligence with respect to this concept if, given an entity (e.g., a horse) exemplifying it, it can generate another entity (e.g., a car) that also exemplifies it, or other novel solutions for transport.

This formulation satisfies the desiderata as follows:
\begin{itemize}
    \item \textbf{Operational:} Intelligence can be measured through conceptual alignment between observable entities and generated entities.
    \item \textbf{Falsifiable:} Systems that consistently generate outputs divergent from the reference concept exhibit low intelligence.
    \item \textbf{Generalizable:} The concept is applicable across humans, biological organisms, and artificial models.
\end{itemize}

Identifying a concept \(k\) whose distinguishers \(F\) cannot separate from the data at hand is itself an act of intelligence.

\section{Interpretations and Special Cases}
\subsection{The Turing Test as a Special Case of $\varepsilon$-Concept Intelligence}
If we define the fibre \(C_k\) (corresponding to a concept \(k\)) as the set of human language responses and let the family of distinguishers \(F\) consist of human judges evaluating whether a given response was produced by a human or a machine, then the classic Turing Test naturally arises as a special case of $\varepsilon$-concept intelligence, with respect to electronic computers and humans.

\subsection{C-test as a Special Case of $\varepsilon$-Concept Intelligence}

Following Hernández-Orallo's C-tests\,\cite{hernandez1998}, we define, for each difficulty level \(h\), the fibre \(C_{k_h}\) (corresponding to a concept \(k_h\)) as the set of all sequences whose Levin complexity equals a fixed value \(h\)
and that admit a single minimal explanation.
Let the distinguisher family \(F\) accept only the continuation produced by this
unique minimal program.  Under this configuration, the
$\varepsilon$-criterion degenerates into a binary verdict: if an agent's
continuation matches the unique minimal explanation, then
$\Delta_{F}^{\sigma}(S,\widehat{S})\le\varepsilon$ and the response is correct; otherwise it
is incorrect.  Hence the classical C-test can be regarded as an
$\varepsilon$-Concept instance with a vanishingly small
$\varepsilon$.

\subsection{Legg--Hutter Intelligence as a Facet of $\varepsilon$-Concept Intelligence}
If we define the fibre \(C_k\) (corresponding to a concept \(k\)) as the set of behaviour sequences that maximise expected reward in a particular environment and let the family of distinguishers \(F\) be the environment’s reward signal judging how closely a generated behaviour approaches that optimum, then, when we consider only an agent interacting with its environment, Legg--Hutter intelligence naturally arises as a facet of $\varepsilon$-concept intelligence.

\section{Foundational Properties of the Definition}
\subsection{Incompleteness and Transformation of Entities}
\textit{The given entities may be incomplete or transformed.} All forms of intelligence can be seen as methods for generating new entities based on given entities. For example, a classic neural network trained on a dataset learns to approximate a surface that locally aligns with the target concept, thereby enabling consistent entity generation. Human intelligence is another method. There will be future machine learning methods that do not rely solely on surface approximation to generate entities.

\subsection{Conceptual Ambiguity of Entities}
\textit{A single set of entities may correspond to multiple possible concepts.} An observable entity can often belong to multiple concepts. The initial entity exemplifying a concept is often generated as an instance of another concept, which can be understood as a form of creativity. Understanding how entities relate to their generative processes—and transitions between concepts—is a profound and ongoing research problem.

\subsection{The Locality of Intelligence}
\textit{Intelligence, as defined here, is local to the given concept.} For example, while a homing pigeon cannot code or speak human language, after just 15--30 days of training it can classify breast-cancer histopathology slides with up to $95\%$ accuracy, and an ensemble of four pigeons achieves an AUC of $0.99$---on par with specialist pathologists\cite{levenson2015pigeon}. Likewise, if a machine consistently translates languages more accurately than a person, it is more intelligent than a person in language translation. More general intelligence would imply the ability to generate entities across a wider range of concepts.

\subsection{Generation as Core Expression of Intelligence}
Our definition emphasizes that intelligence is not merely a matter of abstract mapping or numeric manipulation—as is often the case in formal mathematics—but fundamentally about generation. The act of generating new entities exemplifying a given concept lies at the heart of intelligent behavior.

This generative perspective manifests across domains. In business, for instance, a company consistently generating products under the same brand identity to adapt to evolving markets. In nature, a species reproduces successive generations of offspring. From the standpoint of our framework, both are valid expressions of intelligence.

Viewed through this lens, intelligence, defined as the ability to generate concept-consistent entities, appears to be a principle widely present in natural systems. The field of bio-inspired engineering, where designs mimic biological structures and processes (e.g., shark skin-inspired surfaces for drag reduction, or ant colony optimization for algorithms), can be seen as implicitly leveraging this principle \cite{Benyus1997,Bechert2000,Dorigo2004}. These biological ``designs'' or behaviors represent highly effective solutions within specific environmental or functional concepts. Thus, bio-mimicry can be interpreted as generating entities (structures, behaviors) that are highly consistent with the success criteria of their respective ecological concepts.

The architectural wisdom of I.M. Pei provides an elegant analogy. He once remarked: ``There is always a theme, there is a certain repetition, but they do not seem like repetition—only the endless variety of a simple theme.''\cite{Pei1987} This resonates deeply with our formulation, in which intelligence expresses itself through variety constrained by coherence—a generative consistency within conceptual identity.

\subsection{Intelligence and Consciousness}
There is a common assumption that intelligence necessarily entails consciousness—that the two must co-exist. This paper does not attempt to define consciousness; however, the proposed definition of intelligence does not presuppose it. Intelligence, as defined here, does not require consciousness.

As a counterexample, consider a minimal computer program designed to solve calculus problems through symbolic manipulation or numerical methods. Within the conceptual framework where the fibre \(C_k\) (corresponding to a concept \(k\)) is the set of correctly solved calculus problems, this program can exhibit extremely high $\varepsilon$-concept intelligence, producing outputs (i.e., solutions) that are indistinguishable from valid mathematical results—likely surpassing the performance of most humans in this narrow domain. Yet few would attribute consciousness to such a deterministic mathematical solver, as its algorithmic nature is generally not regarded as a sufficient condition for consciousness.

\subsection{Intelligence and Learning Ability}
It is frequently assumed that intelligence necessitates an inherent capacity for learning. Our proposed definition, however, does not mandate this. We can illustrate this distinction with a counterexample within our framework. Suppose a standard LLM is capable of consistently generating correct answers (\(\widehat{S}\)) for problems belonging to a specific concept \(k\). Now, we fix its internal parameters and restrict its computational process to be purely deterministic. Such a system is clearly incapable of learning, as it cannot update its internal state based on new information or interactions, but according to our definition, it is intelligent within that domain. Clearly, learning ability is related to the change in \(\Delta_{F}^{\sigma}(S, \widehat{S})\) instead.

\subsection{Clarifications on Scope and Interpretation}
\begin{itemize}
    \item \textbf{Classification:} In classification tasks, each input-output pair is an entity. A system that accurately produces such entities, exemplifying a joint concept, aligns with that concept and is thus intelligent by this definition.
    \item \textbf{Decision-Making:} In decision-making or interactive environments, we identify a concept (e.g., ``goal-achieving behaviors'') whose fibre is the set of behaviors (which are entities) likely to achieve a given objective (e.g., survival, success, reward maximization). Producing an entity here corresponds to selecting a behavior consistent with this concept.
    \item \textbf{Memory-Based Methods:} Systems that rely heavily on memorization or exhaustive enumeration can still qualify as intelligent if their outputs remain consistent with the target concept. The significance of memorizing entities from a concept is often underestimated.

    \item \textbf{Relation to Chollet’s Measure of Intelligence:} Our definition focuses on fidelity ($\epsilon$), deliberately separating it from the \textit{efficiency} of achieving that fidelity. This allows us to frame measures like Chollet's~\cite{Chollet2019}, which emphasize the rate of skill acquisition, as assessments of a crucial, distinct dimension of intelligence—what we later term \textit{diachronic capability}. A system can thus be highly intelligent (low $\epsilon$) yet inefficient in learning, or vice versa.

\end{itemize}

\section{A Dynamic Framework for General Intelligence}
\subsection{On the Selection of Concepts and Distinguishers}

The operationalization of our framework hinges on the principled selection of a concept \(k\) (and its corresponding fibre \(C_k\)) and its family of distinguishers \(F\). These choices are governed by pragmatic fitness and measurement resolution, rather than being arbitrary.

First, the selection of a concept \(k\) (and its corresponding fibre \(C_k\)) is not a normative judgement of its intrinsic ``rightness'' or ``goodness.'' Instead, this constitutes a process of pragmatic selection: concepts whose fibres effectively satisfy human needs become more prevalent because of their fitness, and are thus more frequently observed and studied. More broadly, human needs represent merely one possible selective environment. In a more general sense, the prominence of any concept (and its fibre) arises from its fitness within a given environment, be it physical, ecological, or an abstract logical system.

Second, given a concept \(k\) (and its fibre \(C_k\)) and a target intelligence level \(\varepsilon\), the choice of the distinguisher family \(F\) is constrained. Any practical \(F\) has its own intrinsic resolution limit, denoted as \(\varepsilon_f\)---its margin of error in determining membership in \(C_k\). A fundamental prerequisite for a meaningful assessment is that \(\varepsilon_f < \varepsilon\), as an instrument cannot resolve details finer than its own precision. This imposes a critical limitation: if an agent's performance (\(\varepsilon\)) far exceeds the evaluator's capability, such that \(\varepsilon\) is significantly smaller than \(\varepsilon_f\), then \(F\) can no longer reliably verify such high-level performance.

Furthermore, while not a formal requirement of the definition itself, for practical evaluation, it is highly desirable that the distinguishers in \(F\) be computationally tractable (e.g., computable in polynomial time). These considerations ensure that the measurement of intelligence constitutes an effective evaluation framework rather than an abstract exercise.

\subsection{Efficiency, Cost, and Dynamic Adaptation}

In principle, within a system that evolves over time, an ideal general intelligent agent should possess the potential to generate entities for any concept whose fibre can be specified or exemplified. In reality, however, any agent at any single point in time~\(t\) confronts a finite set of concepts (each with its corresponding fibre), say \(K_t\). It is this very transition from infinite potential to finite reality that necessitates evaluating intelligence from a dynamic perspective. From this vantage point, intelligence can be assessed along two dimensions:\footnote{The terms ``synchronic'' and ``diachronic'' are adopted from structural linguistics to distinguish the study of a phenomenon at a particular point in time from its evolution through time, respectively. See Saussure (1916)~\cite{Saussure1916}.}

\begin{enumerate}
    \item \textbf{Synchronic Capability:} An agent's breadth and capacity to generate low-$\varepsilon$ entities for the set of concepts \(K_t\) at a given time~\(t\).
    
    \item \textbf{Diachronic Capability:} The ability of an agent to transition from a state of competence for a set of concepts \(K_t\) to a subsequent set \(K_{t'}\). This essentially measures the agent's adaptive capacity.
\end{enumerate}

Within this framework, concepts such as efficiency, cost, compression, and learning acquire their \textit{instrumental value}, their importance realized through their \textit{potential contribution} to these two capabilities. Whether and to what extent an agent leverages these properties stems from the nature of the challenges it confronts---namely, the scale, complexity, and temporal dynamics of the concept sets \(K_t\).

For instance, when confronted with a large and diverse \(K_t\), a low-cost generation process might be a strategy for enhancing synchronic capability. Likewise, when concept sets change frequently and unpredictably, a highly compressed knowledge representation could become an advantage for boosting diachronic capability. These are not universal laws, however; the ultimate determinant of a strategy's merit remains the agent's overall ability to sustainably generate concept-consistent entities.

\section{Generalization to Unseen Concepts}
\medskip
We now consider a training setup based on multiple concepts. Let the training data consist of entities:
\[
x_{1,1},\,\dots,\,x_{1,m_1}; \quad
x_{2,1},\,\dots,\,x_{2,m_2}; \quad \dots \quad
x_{M,1},\,\dots,\,x_{M,m_M}
\]
where each set
\(\{x_{i,1},\,x_{i,2},\,\dots,\,x_{i,m_i}\}\)
is drawn from a distinct concept \(k_i\) (and its corresponding fibre \(C_{k_i}\)), for \(i=1,2,\dots,M\).
Assume each entity \(x_{i,j}\) can be viewed or transformed into a problem–solution pair, depending on the task.
Now define
\[
S_i \;=\;\{\,x_{i,1},\,x_{i,2},\,\dots,\,x_{i,m_i}\},
\]
so that each \(S_i\) (drawn from \(C_{k_i}\)) represents a single entity from a higher-level training space composed of concepts. By training on multiple such sets \(\{S_1,S_2,\dots,S_M\}\), the model learns to generalize beyond any single concept: it acquires the ability to generate new problem–solution pairs from previously unseen concepts. This supports generalizable intelligence: the ability to produce coherent outputs for unseen problems drawn from concepts not encountered during training.

A practical example of this can be seen in large language models (LLMs). During training, LLMs observe data drawn from many different implicit concepts—spanning languages, domains, and tasks. Although not explicitly trained for each specific case, these models can often generate accurate responses to novel prompts by generalizing across the diverse conceptual patterns they have encountered.

Furthermore, suppose a family of admissible distinguishers and an
\(\varepsilon\)-concept intelligence metric have been established on a reference
concept, which we denote as \(\mathcal{K}_A\).
If an unseen problem space, formalized as another concept \(\mathcal{K}_B\), can be embedded into \(\mathcal{K}_A\) via a functor
\(F:\mathcal{K}_B\!\to\!\mathcal{K}_A\) that is \emph{full}, \emph{faithful},
and \emph{essentially surjective}, then every validated distinguisher in
\(\mathcal{K}_A\) can, in principle, be transported along \(F\) to \(\mathcal{K}_B\)\cite{MacLane1971},
thereby conferring the same theoretical \(\varepsilon\)-concept intelligence bound in the new domain.

\section{Conclusion}
This paper proposes a minimal, testable, and general definition of intelligence grounded in the ability to generate entities consistent with a given concept. It aims to unify a wide range of intelligent behaviors under a single operational framework, without relying on task-specific or anthropocentric assumptions.

First, in terms of evaluation, by leveraging the principle of entity indistinguishability one can quantify intelligence through the degree of alignment between a system’s outputs and given entities on concept attributes. Whether dealing with a generative model, a classifier, or a reinforcement‐learning agent, if its outputs are indistinguishable from target-concept entities within an acceptable error bound, it can be said to possess the corresponding \(\varepsilon\)-concept intelligence.

Second, regarding robustness and safety, this definition underscores that the quality of training data directly determines the safety and reliability of system outputs. By incorporating carefully curated, safety-compliant synthetic data at the initial training stage, harmful outputs can be mitigated at their source—potentially offering a first-mover advantage over purely post-hoc filtering or correction.

Third, the proposed framework naturally supports generalization to novel tasks and unseen concepts. Once the system has learned multiple known concepts, it can leverage its abstract understanding of concept structure to generate consistent entities for new, previously unencountered concepts—thereby achieving genuine generalized intelligence. This property aligns with current large-scale pretraining practices.

A promising direction for future research is to elevate this framework using the formal machinery of Category Theory. The ``concepts'' we discuss could be rigorously defined as objects in a category, where the morphisms capture structural similarities. In such a setting, generalization to new problems could be modeled as a functor mapping one problem concept to another. This would allow for the formal ``transport'' of distinguishers and provide a deeper structural account of generalizable intelligence. Exploring this possibility is a key next step for this work.

Subsequent work develops the concept--fibre--entity perspective introduced here into a broader mathematical framework based on directed similarity fields, concept fibres, and generative operators~\cite{Ng2025SFT}.

By focusing on observable generative behavior, this definition lays the foundation for theoretical development, cross-domain evaluation, and the design of safe and general-purpose intelligent systems.
\begin{credits}
\subsubsection{\discintname}
The authors have no competing interests to declare that are relevant to the content of this article.
\end{credits}

%
%
%

\end{document}